%% file: main.tex
\documentclass[runningheads]{llncs}
\usepackage{graphicx,}
\usepackage{amsmath,dsfont,comment}
\usepackage{amssymb}
\usepackage{color}
\usepackage{xcolor}
\definecolor{myblue}{gray}{.55}
\DeclareMathOperator*{\argmax}{arg\,max} 
\newcommand{\etal}{\textit{et al}.}
\newcommand{\ie}{\textit{i}.\textit{e}.}
\newcommand{\eg}{\textit{e}.\textit{g}.}
\newcommand{\etc}{\textit{etc}.}
\newcommand{\wrt}{\textit{w}.\textit{r}.\textit{t}.}

\newcommand{\fu}[1]{\textcolor{blue}{#1}}

\usepackage[linesnumbered,vlined,ruled]{algorithm2e}
\usepackage{subfigure}
\usepackage{threeparttable}
\usepackage{multirow}
\usepackage{array}
\newcolumntype{L}[1]{>{\raggedright\arraybackslash}p{#1}}
\newcolumntype{C}[1]{>{\centering\arraybackslash}p{#1}}
\newcolumntype{R}[1]{>{\raggedleft\arraybackslash}p{#1}}

\usepackage{url}

\usepackage[pagebackref=true,breaklinks=true,colorlinks,bookmarks=false]{hyperref}

\begin{document}

\mainmatter  

\title{Accelerating Robustness Verification of Deep Neural Networks Guided by Target Labels}


\titlerunning{Accelerating Robustness Verification of Deep Neural Networks}

\author{Wenjie Wan\inst{1} \and Zhaodi Zhang\inst{1} \and Yiwei Zhu\inst{1} \and Min Zhang\inst{1} \and Fu Song\inst{2}}

\authorrunning{
	Wan \etal
}

\institute{
	Shanghai Key Laboratory of Trustworthy Computing,\\
East China Normal University, Shanghai, China\\
	\texttt{\{51184501150,zdzhang,51184501094\}@stu.ecnu.edu.cn,\\ zhangmin@sei.ecnu.edu.cn}
	\and ShanghaiTech University, Shanghai, China\\
	\texttt{songfu@shanghaitech.edu.cn}
}

\maketitle

\input{abstract}

\input{intro}

\input{prelim}

\input{approach}

\input{expri}

\input{conc}

\end{document}

%% file: abstract.tex
\begin{abstract}
Deep Neural Networks (DNNs) have become key components of many safety-critical applications such as autonomous driving and medical diagnosis.
However, DNNs have been shown suffering from poor robustness because of their susceptibility to adversarial examples
such that small perturbations to an input result in misprediction.
%
Addressing to this concern, various approaches have been proposed to formally verify the robustness of DNNs.
Most of these approaches reduce the verification problem to optimization problems of searching an adversarial example for a given input so that
it is not correctly classified to the original label.
However, they are limited in accuracy and scalability.
In this paper, we propose a novel approach that can accelerate the robustness verification techniques  by 
guiding the verification with target labels. The key insight of our approach is that
the robustness verification problem of DNNs can be solved by verifying sub-problems of DNNs,
one per target label. Fixing the target label during verification can drastically reduce the search space and thus improve the efficiency.
We also propose an approach by leveraging symbolic
interval propagation and linear relaxation techniques to sort the target labels in terms of chances that adversarial examples exist.
This often allows us to quickly falsify the robustness of DNNs and the verification for remaining target labels could be avoided.
Our approach is orthogonal to, and can be integrated with, many existing verification techniques.
For evaluation purposes, we integrate it with three recent promising DNN verification tools, \ie, \texttt{MipVerify}, \texttt{DeepZ}, and \texttt{Neurify}.
Experimental results show that our approach can
significantly improve these tools by 36X speedup
when the perturbation distance is set in a reasonable range.


\end{abstract}

%% file: intro.tex
\section{Introduction}
Deep Neural Networks (DNNs) have achieved remarkable performance and accomplished unprecedented breakthrough in many complex tasks such as image classification~\cite{krizhevsky2012imagenet,he2016deep} and speech recognition~\cite{hinton2012deep}. The progress makes it possible to apply DNNs to real-world safety-critical domains, \eg, autonomous driving~\cite{Holley18,Apollo,Waymo} and medical diagnostics~\cite{CGGS12,SWS17,PCG15}.
Systems in such domains must be highly dependable and hereby their safety should be comprehensively certified before deployments.
One of the most challenging problems in this domain is that DNNs have been shown suffering from poor robustness. That is, a small modification to a valid input may cause systems to make completely wrong decisions~\cite{szegedy2013intriguing,goodfellow2014explaining,moosavi2016deepfool,LCFSL20,CCFDZSL20,DZBS19}, which consequently result in serious consequences and even disasters. For instance, a Tesla car in autopilot mode caused a fatal crash as it failed to detect a white truck against a bright sky with white clouds~\cite{stewart2018tesla}.
Therefore, it is important and necessary to certify the robustness of DNN-based systems before deployments by proving that the neural networks can always make the same prediction for a valid input even if the input is slightly perturbed within an allowed range due to uncertainties from the environment or adversarial attacks.


Many efforts have been made to certify the robustness of DNNs using formal verification techniques~\cite{carlini2017towards,katz2017reluplex,peck2017lower,hein2017formal,dvijotham2018dual,weng2018towards,zhang2018efficient,gehr2018ai2,singh2018fast,tjeng2017evaluating,ehlers2017formal,singh2019abstract,singh2018boosting,wang2018efficient}. The essence of certifying the robustness is to \textit{prove} mathematically the absence of adversarial examples for a DNN within a range of allowable perturbations, which are usually provided by a valid input and a $L\text{-}norm$ distance threshold.
There are three main criteria of evaluating verification approaches: \textit{soundness}, \textit{completeness} and \textit{scalability}.
The first states that if a DNN passes the verification, then there are no adversarial examples.
The second states that every robust DNN should pass the verification.
The last one indicates the scale of DNNs that a verification method can handle. It is known that the verification problem of DNNs with Rectified Linear Unit (ReLU) activation function is NP-complete~\cite{katz2017reluplex}. This means that sound and complete verification approaches usually have limited scalability.
Existing formal verification approaches  either have limited scalability and can only handle small networks~\cite{katz2017reluplex,lomuscio2017approach,cheng2017maximum,fischetti2017deep},
or rely on abstraction techniques that simplify the verification problem for better scalability, but they may produce false positives~\cite{gehr2018ai2,singh2018fast,singh2019abstract} after loosing the completeness property due to the introduction of abstraction.

\noindent
{\bf Our contribution}.
In this work, we propose a generic approach that can enhance neural network verification techniques
by guiding the verification with target labels ---- thus
making it more amenable to verification.
Our approach is based on the following key insights. Many existing approaches reduce the verification
problem to some optimization problems of searching an adversarial example for a given input so that it is not correctly classified to the original
label. We found that by fixing a target label during
verification, the search space could be drastically reduced so that
the verification problem with respect to the target label can be efficiently solved, while the overall verification problem
can be solved by verifying the DNN for all the possible target labels. Specifically,
guided by the target label, we can efficiently compute an adversarial example if there exists one for the given input and $L\text{-}norm$ distance threshold $\epsilon$.
In this case, the robustness of the DNN is falsified and the verification for other target labels can be avoided.
Furthermore, rather than choosing target labels randomly, we propose an algorithmically efficient approach to sort the target labels by leveraging the \emph{symbolic interval propagation} and \emph{linear relaxation} techniques, so that the target labels to which some inputs are misclassified by the DNN with larger probabilities are processed first.
This often allows us to quickly disprove the robustness of the DNN when the target DNN is not robust.


Our approach is orthogonal to, and can be integrated with, many existing verification techniques which are leveraged to verify
the robustness of DNNs for target labels. To evaluate the effectiveness and efficiency of our approach, we integrate it with three recent promising neural network verification tools, \ie\ \texttt{MipVerify}~\cite{tjeng2017evaluating}, \texttt{DeepZ}~\cite{singh2018fast}, and \texttt{Neurify}~\cite{wang2018efficient}. We compare both the verification result and the time cost for verification of the original tools and the tools integrated with our approach. Experimental results show that  our approach can help the three tools achieve up to 36X acceleration in time efficiency under reasonable perturbation thresholds.
Furthermore, the properties \ie\ \emph{soundness} and \emph{completeness} (if satisfied) of the original tools are still preserved.



In summary, this paper makes the following three main contributions:
\begin{itemize}
\item A novel, generic approach for accelerating the robustness verification of neural networks guided by target labels.


\item An approach for sorting target labels by leveraging the symbolic interval propagation and linear relaxation techniques.

\item Extensions of three recent promising neural network verification tools with the proposed approach. 


\end{itemize}

\noindent
\textbf{Outline.} Section~\ref{sec:prelim} briefly introduces some preliminaries used in this work.
Section~\ref{sec:approach} presents our verification approach.
Section~\ref{sec:expri} reports experimental results.
Section~\ref{sec:relatedwork} discusses related work.
Section~\ref{sec:conc} finally concludes the paper and discusses some future work.

%% file: prelim.tex
\section{Preliminaries}\label{sec:prelim}
In this section, we recap some preliminaries such as feed-forward deep neural networks, interval analysis, symbolic interval propagation and linear relaxation that are necessary to understand our approach.

\subsection{Feed-Forward Deep Neural Networks}
In this work, we consider feed-forward deep neural networks (FNNs).
An $l$-layer FNN can be considered as a function $f:I\rightarrow O$, mapping the set of vectors $I$ to the set of vectors $O$. 
Function $f$ is recursively defined as follows: 
\begin{equation}\label{def:fnn}
\begin{aligned}
x^{0} &= x,\\
x^{k+1}& = \phi(W^{k}x^{k} + b^{k})  \quad \mbox{for} \  k = 0, ..., l-1, \\
f(x) &= W^{l}x^{l} + b^{l}.
\end{aligned}
\end{equation}
where $x^{0} = x \in I$ is the input, $W^{k}$ and $b^{k}$ respectively are the weight matrix and bias vector of the $k$-th layer,
and $\phi(\cdot)$ (e.g., {\tt ReLU}, {\tt sigmoid}, {\tt tanh} \etc) is an activation function applied coordinate-wise to the input vector.
{\tt ReLU}, defined by $\texttt{Relu}(x)\equiv\max(0,x)$, is one of the most popular activation functions used in the modern state-of-the-art DNN architectures~\cite{he2016deep,huang2017densely,szegedy2016rethinking}. In this paper we are focused on FNNs that only take \texttt{ReLU} as the activation function.
For a given input $x$, the \emph{label} of $x$ is determined by the function $\mathcal{L}$, defined as, \[\mathcal{L}(f(x))=\argmax_{j} f(x)[j],\]
where, $f(x)[j]$ denotes the $j$-th element in the output vector $f(x)$ which is the confidence that $x$ is classified to the label $j$.
In the case that the last step is not well defined,
namely, there are more than one maximum elements in $f(x)$,  we call that $x$ admits an adversarial
example. Hereafter, we assume that the last step of an FNN is well defined, otherwise it is not robust.
By applying the \emph{softmax} function to the output $f(x)$, we will get the probabilities of the labels to which the input $x$ is classified.
For this reason, we in what follows may say $f(x)[j]$ is the probability that the input $x$ is classified to the label $j$.
For simplicity, we also use the indices $j$ to represent the classification labels. $\mathcal{L}(f(x))$ returns the label whose corresponding probability is the largest among all the labels.
We call it original label of the input $x$.

\begin{definition}[Robustness of FNNs]
Given an FNN $f:I\rightarrow O$, an input $x\in I$, and an $L_{norm}$ distance threshold $\epsilon$, $f$ is \emph{robust} \wrt\ $x$ and $\epsilon$ if
\[\mathcal{L}(f(x))=\mathcal{L}(f(x'))\]
for all $x'\in I$ such that $L_{norm}(x,x')\leq \epsilon$.

If there exists some $x'\in I$ such that $\mathcal{L}(f(x))\neq\mathcal{L}(f(x'))$, $x'$ is called an \emph{adversarial example} of $x$.

Given a target label $j$ such that $j\neq \mathcal{L}(f(x))$, the FNN $f$ is called \emph{j-robust} \wrt\ $x$ and $\epsilon$, if
$f(x')[j]< f(x')[j']$ for all $x'\in I$ such that $L_{norm}(x,x')\leq \epsilon$, where $j'$ denotes the original label $\mathcal{L}(f(x))$ of $x$.
\end{definition}

The next proposition states that the robustness problem of a DNN can be reduced to the $j$-robustness problem of the DNN (which up to
our knowledge has never been stated in the literature though straightforward):

\begin{proposition}\label{prop:robustness}
Given an FNN $f:I\rightarrow O$, an input $x\in I$, and an $L_{norm}$ distance threshold $\epsilon$, suppose
$J$ is the set of all the possible labels of $f$, then:

\begin{center}
\begin{tabular}{c}
$f$ is robust \wrt\ $x$ and $\epsilon$ \\
 iff \\
$f$ is j-robust \wrt\ $x$ and $\epsilon$,
 for all $j\in J\setminus\{\mathcal{L}(f(x))\}$.
\end{tabular}
\end{center}

\end{proposition}

In this work, we only consider $L_\infty$ norm, that is: for each pair of vectors $x,x'$ with the same size,
$$L_\infty(x,x')\equiv\max\{\left|x[j]-x'[j]\right|  \ : \  j \ \mbox{is an index of the vector} \ x \}.$$

\subsection{Interval Analysis}

Interval analysis is a technique which works on intervals rather than concrete values, where an interval represents
a set of consecutive, concrete values. We provide some basic terms, concepts, and operations of intervals below.

An \emph{interval} $X$ is a pair $[\underline{X},\overline{X}]$, where $\underline{X}$ is the lower bound, and $\overline{X}$ such that $\overline{X}\geq \underline{X}$ is the upper bound.
The interval $[\underline{X},\overline{X}]$ denotes the set of concrete values $\{i\in  \mathds{N}\mid \underline{X}\leq i\leq \overline{X}\}$.

The basic arithmetic operations between intervals are defined in~\cite{moore2009introduction}. In this paper, we only present the definitions of the addition, difference and scalar multiplication operations which are sufficient for this work.
The key point of these definitions is that computing with intervals is computing with sets. By definition,  the addition ($+$) of two intervals $X$ and $Y$ is the set
$$X + Y = \{x+y : x \in X, y\in Y\}=[\underline{X}+\underline{Y}, \overline{X}+\overline{Y}].$$
For example, let $X = [0,2]$ and $Y = [-1,1]$. Then $X+Y = [0 + (-1), 2 + 1] = [-1, 3]$. The {difference} ($-$) of two intervals $X$ and $Y$ is the set denoted by $X-Y$, which is defined as follows:
$$X - Y = \{x-y : x \in X, y\in Y\}=[\underline{X}-\overline{Y}, \overline{X}-\underline{Y}].$$
For instance, let $X = [-1,0]$ and $Y = [1,2]$. Then we have $-Y$ = [-2,-1] and $X-Y = X+(-Y)=[-3, -1]$.
The scalar multiplication ($\cdot$) between an interval $X$ and a constant $c$ is the set, denoted by
$c\cdot X$ or $c X$, which is defined as follows:
$$c\cdot X= \{c\times x : x \in X\}=[c\times\underline{X}, c\times\overline{X}].$$
For instance, let $X$ = [-1,3] and $c=2$.  Then, we have $c\cdot X=[-2,6]$.

\subsection{Symbolic Interval Propagation}
To sort target labels for  a DNN $f:I\rightarrow O$, an input $x\in I$ and a distance threshold $\epsilon$,
we will propagate the interval from the input layer to the output layer
via interval propagation.  However, naively computing the output interval of the DNN in this way
suffers from high errors as it computes extremely loose bounds due to the dependency problem.
In particular, it may get a very conservative estimation of the output,
which is not tight enough to be useful for sorting labels.



\begin{figure}[!h]
\begin{center}
   \subfigure[\scriptsize Naive interval propagation]{
    \label{interval_propagation}
    \includegraphics[width=0.36\textwidth]{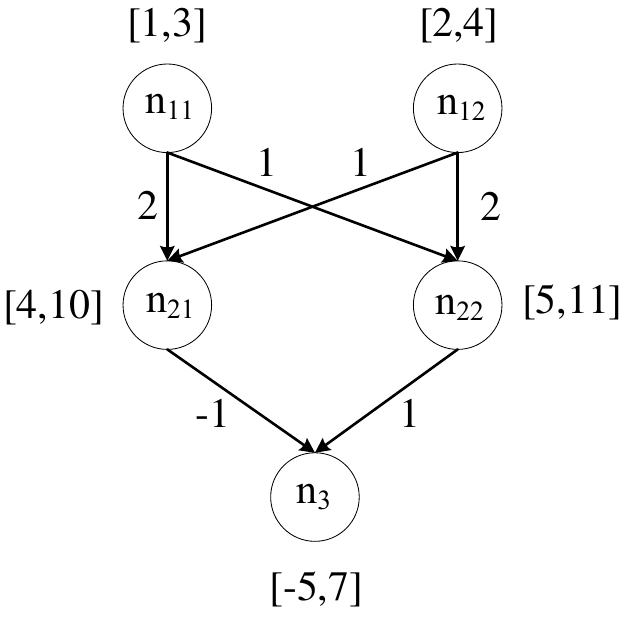}
  }
\hspace{2mm}
  \subfigure[\scriptsize Symbolic interval propagation]{
    \label{symbolic_propagation}
    \includegraphics[width=0.36\textwidth]{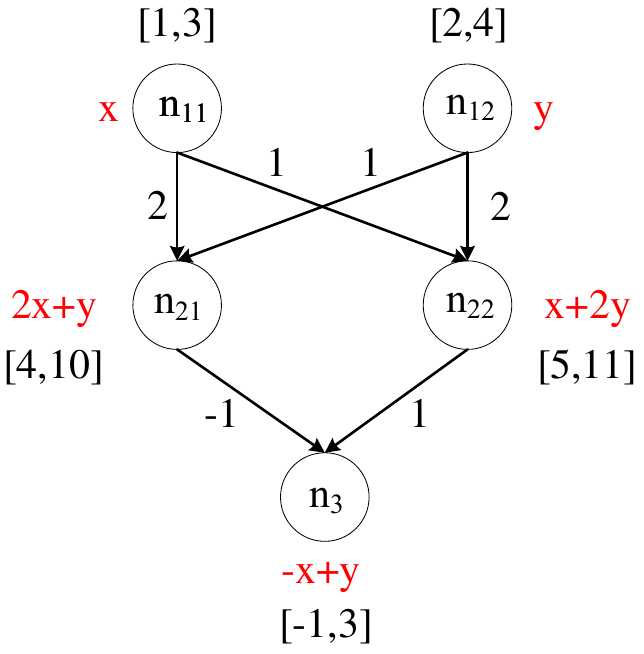}
  }
\vspace{-2mm}
   \caption{Naive interval propagation vs. symbolic interval propagation.}
\label{propagation}
\end{center}
\vspace{-15pt}
\end{figure}
Consider a 3-layer DNN given in Figure~\ref{interval_propagation}, where the weights are associated to the edges and all elements of the bias vectors are $0$.
Suppose the input of the first layer are the intervals $[1,3]$ and $[2,4]$.
By performing the scalar multiplications and additions over intervals layer-by-layer, we get the output interval $[-5,7]$.
This output interval contains the concrete value $5$ which is introduced by overestimation, but is an infeasible value.
For instance, $-5$ can occur only when the neuron $n_{21}$ outputs $10$ and the neuron $n_{22}$ outputs $5$.
To output $10$ for the neuron $n_{21}$,
the neurons $n_{11}$ and $n_{12}$ should output $3$ and $4$ simultaneously.
But, to output $5$ for the neuron $n_{22}$,
the neurons $n_{11}$ and $n_{12}$ should output $1$ and $2$ simultaneously. This effect is known as the \emph{dependency problem}~\cite{moore2009introduction}.


Symbolic interval propagation~\cite{WangPWYJ18} is a technique to minimize overestimation of outputs
by preserving as much dependency information as possible during propagating the intervals layer-by-layer.
A \emph{symbolic interval} is a pair of linear expressions $[e,e']$ such that $e$ and $e'$ are defined over the input
variables.

Let us consider the same example using symbolic interval propagation as shown in Figure~\ref{symbolic_propagation}.
Suppose $x$ and $y$ are the input variables of the neurons $n_{11}$ and $n_{12}$.
By applying the linear transformation of the first layer, the values of the neurons $n_{21}$ and $n_{22}$ are $2x+y$ and $x+2y$ respectively.
Since $x\in \left [ 1,3 \right ]$ and $y\in \left [ 2,4 \right ]$, we have: $2x+y>0$ and $x+2y>0$.
Therefore, the output symbolic intervals of the neurons $n_{21}$ and $n_{22}$ are
$[2x+y,2x+y]$ and $[x+2y,x+2y]$ respectively. By applying the linear transformation of the second layer,
the value of the neuron $n_3$ is $-x+y$. Thus, the output of the DNN will be $\left [ -x+y, -x+y \right ]$.
From $x\in \left [ 1,3 \right ]$ and $y\in \left [ 2,4 \right ]$, we can conclude that the output interval of the DNN
is $\left [-1,3\right ]$, which is tighter than the interval $[-5,7]$ produced by directly performing interval propagation.



This example shows that symbolic interval propagation characterizes how each neuron computes results in terms of the symbolic intervals and related activation functions. As the symbolic intervals keep the inter-dependence between variables, symbolic interval propagation significantly reduces the overestimation.

\subsection{Linear Relaxation}\label{subsec:rel}
To tackle the non-linear activation function \texttt{ReLU}, we use \emph{linear relaxation}~\cite{zhang2018efficient} to strictly overapproximate the symbolic intervals.
Consider an intermediate node with $n=\texttt{ReLU}(X)$.
For each symbolic interval $X=(l, u)$,
based on the signs of $l$ and $u$ (determined by concretizing the symbolic intervals), we consider three cases as shown in Table~\ref{upper/lower_bound}.

\begin{table}[t]
	\caption{Interval propagation for  \texttt{ReLU}.}
	\label{upper/lower_bound} \vspace{-5pt}
	\centering
	\begin{tabular}{cccc}
		\hline
		\centering \textbf{Activation Function} & \multicolumn{3}{c}{\textbf{ReLU}} \\
		\hline
		Condition & Upper & & Lower \\
		\hline
		$0<l<u$ & $x$ & &  $x$ \\
		\hline
		\multirow{2}[2]{*}{$l<0<u$} & $a(x-l)$           &            & $ax$                     \\
		& w.r.t. $(\frac{u}{u-l} \leq a\leq 1)$    &  $~~~~$   & w.r.t. $(0\leq a \leq 1)$       \\
		\hline
		$l<u<0$ & $0$ & &  $0$ \\
		\hline
	\end{tabular}
\vspace{-5mm}
\end{table}

\begin{figure}[!h]
	\centerline{\includegraphics[height=2.2in]{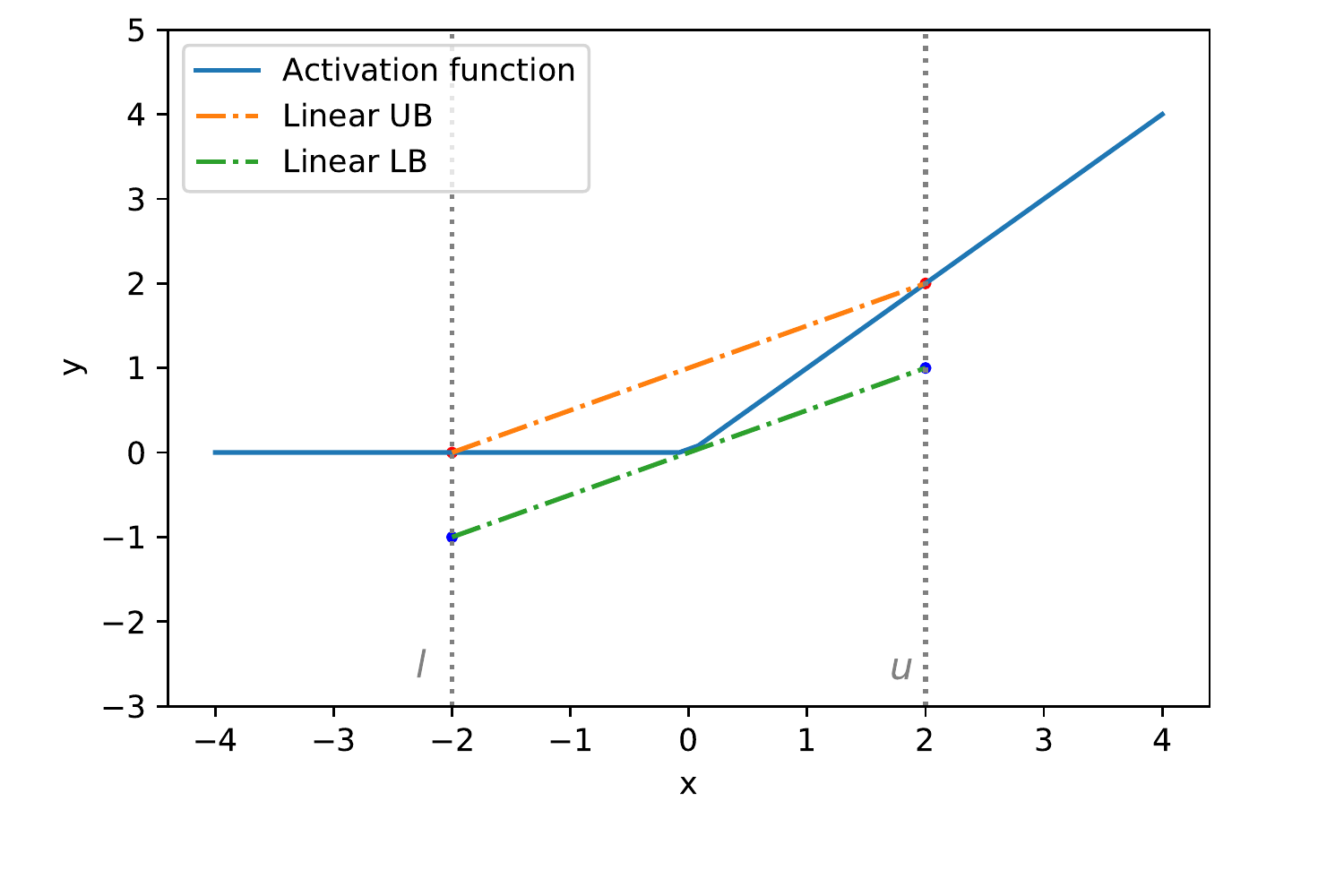}}\vspace{-25pt}
	\caption{The linear upper and lower  bounds for \texttt{ReLU} when $l<0<u$.} \label{relu_middle}
		\vspace{-30pt}
\end{figure}

\begin{itemize}
  \item If $(l>0)$, then $\texttt{ReLU}(x) = x$ for every $x\geq l$.
Thus, $\texttt{ReLU}(X)$ is $X$.
  \item If $(u\leq 0)$, then $\texttt{ReLU}(x) = 0$ for every $x\leq u$.
Thus, $\texttt{ReLU}(X)$ is $[0,0]$.
  \item If $(l\leq 0\leq u)$, then $X$ contains both positive and negative values. The output cannot be exactly represented by one linear interval and thus relaxation is required.
We adopt the linear relaxation defined in~\cite{zhang2018efficient}.
As shown in Figure~\ref{relu_middle}, we can set upper bound as $a(x-l)$ with respect to $\frac{u}{u-l} \leq a\leq 1$ and lower bound as $ax$ with respect to $0\leq a\leq 1$.
We select the value of $a$ to minimize the overestimation error introduced by the linear relaxation.
\end{itemize}
\vspace{-10mm}

%% file: approach.tex
\section{Verification Framework}\label{sec:approach}



In this section, we first show how to sort target labels by leveraging the symbolic interval propagation and linear
relaxation techniques and then present our target label guided verification approach.


\begin{figure}
	\begin{center}
		\includegraphics[width=\textwidth]{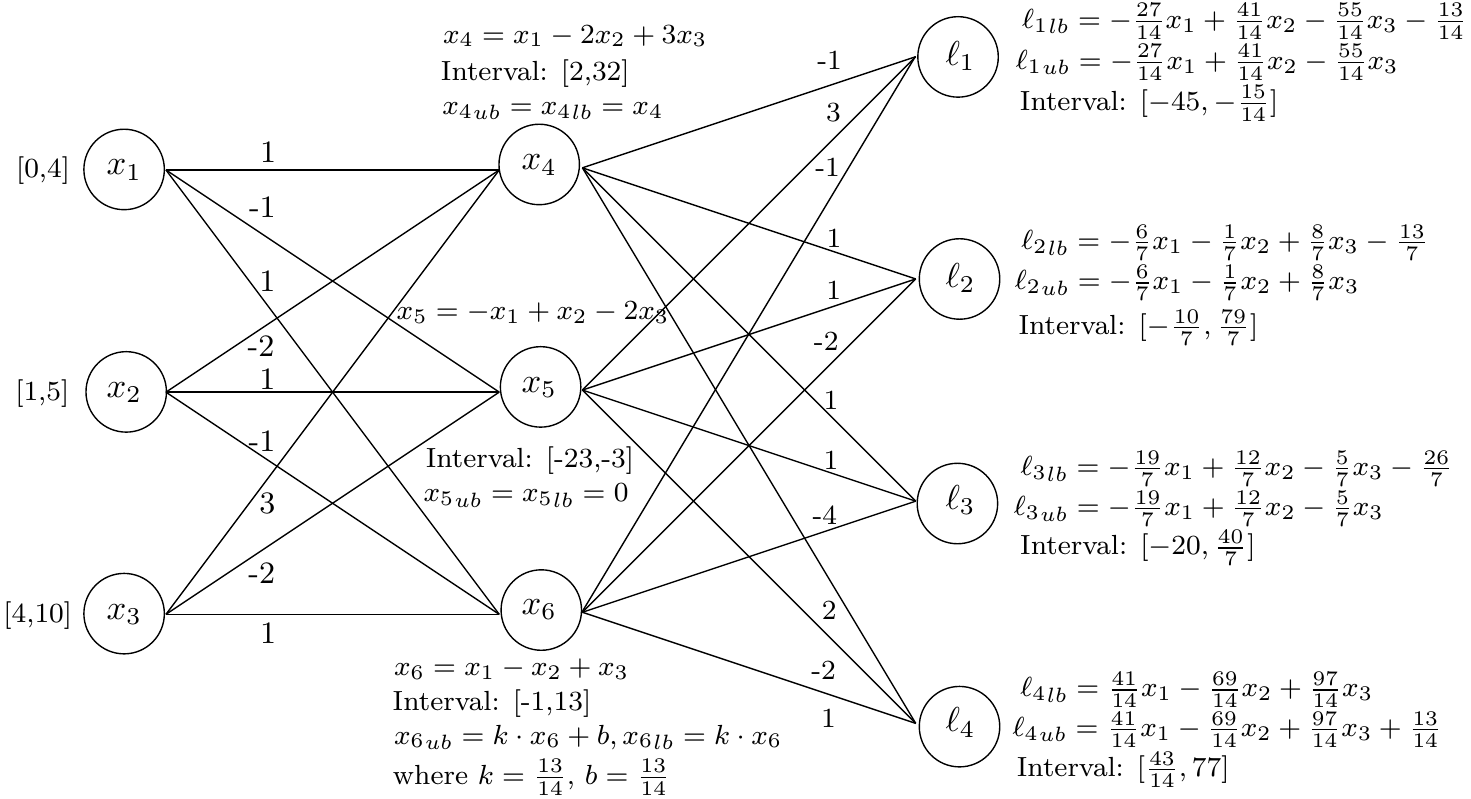}
		\caption{An example of sorting target labels based on output interval}
		\label{fig:sorting}
		\vspace{-1cm}
	\end{center}
\end{figure}

\subsection{Sorting Target Labels} \label{subsec:loc}
To the best of our knowledge, most existing approaches to the robust verification problem for a given FNN and an input
reduce to some other problems, which checks whether adversarial examples in the range that are not correctly classified to the original label of the input exist or not.
Therefore, the search space is relatively larger. Instead of considering all the other labels in one verification, we propose to sort labels and verify FNNs
for each label in order to reduce the search space.

A premise order of labels is that the larger that the probability of a label has,
the larger that the probability of finding an adversarial example for this label, thus,
the higher priority that the label should be verified. A na{\"i}ve approach for label sorting is to calculate the probabilities of all the labels of a target FNN for the given input $x$, as the representative of all the possible inputs.
Though feasible, the sorting result also relies on the distance threshold $\epsilon$, this intuitive approach may lead to imprecise sorting result,  which consequently misleads
the following-up verification.

When considering the distance threshold $\epsilon$, it is infeasible if not impossible to compute the probabilities of all the labels of a target FNN by enumerating all the possible inputs $x'\in I$ that satisfies $L_{\infty}(x,x')\leq \epsilon$. To address this technical challenge, we propose a novel approach by leveraging the symbolic interval propagation and linear relaxation techniques.

Given an FNN $f: I \rightarrow O$, for every $x\in I$ and $L_\infty$ distance threshold $\epsilon$,
we approximate the output range (i.e., output interval) of the FNN for the input $x$ and the distance threshold $\epsilon$,
by leveraging the symbolic interval propagation and linear relaxation techniques.
Firstly, we encode
the set of all the possible inputs $x'\in I$ such that $L_\infty(x,x')\leq \epsilon$ as an input interval.
By applying the symbolic interval propagation with the linear relaxation to handle the \texttt{ReLU} function,
we obtain the output interval.  Finally, we can approximate the probabilities of all the labels of all the possible inputs based on the approximated output range.
The labels except for the original label of the input $x$ are then sorted
in the descending order of the probabilities of the labels for all the possible inputs.

Figure~\ref{fig:sorting} shows an example of computing the output interval for label sorting using symbolic interval propagation.
For the node $x_6$ whose interval is [-1,13], we use linear relaxation to represent the upper bound ${x_6}_{ub}$ and ${x_6}_{lb}$ by two linear constraints.
Assume that $\ell_4$ is the original label of some input. Then, $\ell_2$ is the most likely target label to which the input with some perturbation would be classified 
because the upper bound of its output interval is larger than that of the other labels except the original label $\ell_4$. 
The input with any perturbation in the range would never be classified to the label $\ell_1$ because its upper bound is less than the lower bound of the original label $\ell_4$. 
Thus, $\ell_1$ can be safely discarded from verification, and the sorting result is $\ell_2; \ell_3$.


The key difference between our approach and the na{\"i}ve one is that: our approach returns an interval of probabilities for each label,
while the latter returns a concrete probability for the label.
The interval of probabilities over-approximates all the possible probabilities for all possible inputs in the input interval.
In contrast, the concrete probability only reflects the classification result of one concrete input.
It is known that symbolic interval propagation does not exclude labels that may cause mispredication~\cite{wang2018efficient},
thus the sorted list of labels produced by our approach consists of all possible target labels and is more likely to be a real one than the one produced by the na{\"i}ve approach.

\subsection{Target Label-Guided Verification}
The overview of our verification approach guided by the sorted labels is shown in Algorithm~\ref{alg:algorithm}.
We first sort the labels as aforementioned (lines 1--2) and
then verify the robustness of the given neural network against the labels one by one in $J'$ \wrt\ the input $x$ and a perturbation distance threshold $\epsilon$ (while-loop).

\begin{algorithm}[t]
	\footnotesize
	\SetKwData{Left}{left}\SetKwData{This}{this}\SetKwData{Up}{up}
	\SetKwFunction{Union}{Union}\SetKwFunction{FindCompress}{FindCompress}
	\SetKwInOut{Input}{input}\SetKwInOut{Output}{output}
	\SetKwProg{Fn}{Function}{ is}{end}
	\caption{Robustness verification of FNN guided by Target Labels}
	\label{alg:algorithm}
	\SetAlgoLined \SetAlgoNoEnd
	\Input{An FNN $f$, an input vector $x$, a distance threshold $\epsilon$}
	\Output{\{Robust, Non-robust with an adversarial example, Unknown\}}
	\BlankLine
	$J:=\texttt{Labels}(F)/\{\mathcal{L}(F(x))\}$  \tcp{\textsl{\textcolor{myblue}{$J$: The list of target labels.}}}
	$J'$:=\texttt{sort}($F,x,\epsilon,J$);  \tcp{\textsl{\textcolor{myblue}{Sort the list $J$ according to the probabilities of the labels to which $x$ with $L_\infty$ distance $\epsilon$ are classified.}}}
	flag:=false; \tcp{\textsl{\textcolor{myblue}{to indicate the unkown case.}}}
	\While{$J'\neq nil$}{
		$j:=\texttt{head}(J')$;  \tcp{\textsl{\textcolor{myblue}{The label $j$ in $J'$ with the largest probability.}}}
		Result:=\texttt{Verifier}($F,x,\epsilon,j$);\ \tcp{\textsl{\textcolor{myblue}{To invoke existing NN verifiers}}}
		\Switch{Result}{
			\Case{true}{
				$J':=\texttt{tail}(J')$;  \tcp{\textsl{\textcolor{myblue}{Remove the head of $J'$}}}
				\textbf{continue}\;}
			\Case{false}{
				$x'$:={\it getAdvExample}($F,x,\epsilon,j$); \tcp{\textsl{\textcolor{myblue}{get  adversarial example}}} 	
				\Return $x'$ \;}
			\Case{unkown}{
				flag:=true;\ \tcp{\textsl{\textcolor{myblue}{fail for the current label, try next one}}}
				$J':=\texttt{tail}(J')$;  \tcp{\textsl{\textcolor{myblue}{Remove the head of $J'$}}}
				\textbf{continue}\;
			}
		}
    }
	\If{flag}{
		\Return unknown; \tcp{\textsl{\textcolor{myblue}{if the tool fails for some label.}}}
	}\Else{
		\Return robust; \tcp{\textsl{\textcolor{myblue}{If $F$ is robust against all the labels in $J'$}}}
	}
		
\end{algorithm}

Let $J'$ denote the sorted list of the labels (line 2), and
$j$ denote the head of $J'$ (line 5), the label to which some adversarial examples are most likely misclassified (Line 5). We verify if the given FNN $f$ is robust or not against the label $j$ by invoking
an oracle $\texttt{Verifier}$ (line 6).  The oracle $\texttt{Verifier}$ takes the FNN $f$, the input $x$, the distance threshold $\epsilon$ and the target label $j$ as inputs, and
outputs \emph{true}, \emph{false} or \emph{unknown}.
There exist several state-of-the-art FNN verification tools for verifying robustness, therefore, instead of developing our own tool for verifying $j$-robustness,
we in this work leverage existing tools to achieve this goal.
 The difference is that we provide the tools with the most likely target label $j$  as additional information besides $F$, $x$ and $\epsilon$.
If the oracle $\texttt{Verifier}$ returns \emph{true} (\ie, robust), the algorithm proceeds to verify the remaining labels.

There are two possible outcomes if the FNN $f$ is not robust against the label $j$, depending on the precision of invoked verification tool $\texttt{Verifier}$.
If the tool is both sound and complete, \eg\  {\sf MipVerify}, it returns a real adversarial example to falsify the robustness of the FNN $f$,
namely, the adversarial example is misclassified to the label $j$.  If the tool is incomplete, \eg\ {\sf DeepZ} and {\sf Neurify}, it may return {\it unknown} after several iteration of refinements.
In this case, we set a \emph{flag} to record this failure and skip this label. After all the labels have been verified without returning any
adversarial examples, the algorithm returns \emph{robust} if \emph{flag} is not true, and \emph{unknown} otherwise.

We remark that the soundness and completeness of our algorithm rely on the oracle $\texttt{Verifier}$ employed in the algorithm.
We assume that the implementation of the oracle $\texttt{Verifier}$ is sound, which is reasonable according to the survey~\cite{LiuALBK19}.
Then, our algorithm is also sound.
By the definition of soundness, if our algorithm returns \emph{robust}, then the FNN $F$ must be robust \wrt\ $x$ and $\epsilon$.
That is straightforward because the oracle $\texttt{Verifier}$ returns true for all labels in $J'$. 
Likewise, we can show that our algorithm is complete if and only if  the implementation of the oracle $\texttt{Verifier}$  is complete.

%% file: expri.tex
\section{Implementation and Evaluation}\label{sec:expri}

\subsection{Implementation}
We implemented Algorithm~\ref{alg:algorithm} using Julia programming language~\cite{bezanson2017julia}.
To evaluate its performance, we choose three recent promising DNN verification tools, \texttt{MipVerify}, \texttt{DeepZ} and \texttt{Neurify},
as backend verifier to verify $j$-robustness for each label $j$.

\begin{itemize}

 \item {\sf MipVerify}~\cite{tjeng2017evaluating} formulates the robustness verification of piecewise-linear neural networks as a mixed-integer program. It improves existing Mixed Integer Linear
Programing (MILP) based approaches via a tighter formulation for non-linearities and a novel presolve algorithm that makes full use of all information available. {\sf MipVerify} is both sound and complete.  However, the underlying approach of {\sf MipVerify} relies on applying linear programming per neuron to obtain tight bounds for the MILP solver, which does not scale to larger networks.
\medskip

    \item {\sf DeepZ}~\cite{singh2018fast} makes use of the abstract interpretation technique and uses the abstract domain {\sf ZonoType} that combines floating point polyhedra with intervals, coupled with abstract transformers for common neural network functions such as affine transforms, ReLU, sigmoid and tanh activation functions, and the maxpool operator. These abstract transformers enable {\sf DeepZ} to efficiently handle both feed-forward and convolutional networks. In contrast to {\sf MipVerify}, {\sf DeepZ} is not complete due to the abstraction of original models. 

   \medskip

\item {\sf Neurify}~\cite{wang2018efficient} uses
symbolic interval analysis and linear relaxation to compute tighter bounds on the network outputs.
Neurify is complete but the linear relaxation method used in Neurify is not complete.
Therefore, it may produce spurious adversarial examples and introduce a directed constraint refinement to deal with spurious adversarial examples by iteratively minimizing the errors introduced during the linear relaxation process.
Neurify is complete and at most requires n steps of refinements (n is the number of cross-0 hidden nodes in the network). 
However, in practice, the refinement process might take too long and thus Neurify sets a time threshold to decide when to terminate.

\end{itemize}

We basically treat the three tools as black boxes in our verification framework.
However, we slightly modified them so that they can accept a target label as its an extra input,
besides a neural network, an input of the neural network and an $L_\infty$ distance threshold.
They call their builtin verification algorithms to verify the robustness against the given label instead of all the possible labels.
We use ${\sf MipVerify}^*$,
 ${\sf DeepZ}^*$, and ${\sf Neurify}^*$ to represent the new tools extended with our approach, respectively.

\subsection{Experimental setting}
To evaluate the effectiveness and efficiency of our approach,  
we compare both the verification precision and execution time of the original tools and their corresponding extensions by our approach, respectively.

\smallskip
\noindent\textbf{Benchmarks.} We use the widely-tested dataset, MNIST~\cite{lecun1998gradient},
which is a dataset of handwritten digits, in grayscale with 28 $\times$ 28 pixels.
The dataset consists of a training set of 60,000 examples, and a test set of 10,000 examples, associated with a label from 10 classes.
We selected the first 100 images from the test set of MNIST
for robustness verification.


\smallskip
\noindent
 \textbf{Architectures.} We use three different architectures of fully connected feedforward networks: 2$\times$24 (FNN\_1), 2$\times$100 (FNN\_2), and 5$\times$100 (FNN\_3),
 where $l\times n$ denotes that the network has $l$ layers and each layer consists of $n$ neurons.
The network FNN\_1 is taken from \texttt{Neurify}~\cite{wang2018efficient},
and the networks  FNN\_2 and the FNN\_3 are taken from \texttt{DeepZ}~\cite{singh2018fast}.
All of them have been pre-trained without adversarial training.


\smallskip
\noindent
 \textbf{Experimental setup.} All the experiments were conducted on a Linux server running Ubuntu 18.04.3 with 32 cores AMD Ryzen Threadripper 3970X CPU @ 3.7GHz and 128 GB of main memory. We set three hour as timeout threshold per execution for all the experiments. For each FNN, we evaluate the performance of the tools under different distance thresholds on the same set of inputs.

\subsection{Evaluation}
We evaluate our approach in terms of verification time and verification precision.
Specifically, for each tool $T$, we denote by $T$ the verification time  of the original tool and $T^*$  the verification time of the corresponding tool extended with our approach.
We also record the execution time for sorting all target labels and denote it by $t_{\mathit{sort}}$.
The total time cost by the extended tools is the sum of the verification time and sorting time.
 We calculate the time reduction rate by $(T-T^*-t_{\mathit{sort}})/T$.
We use the form $m/n$ to represent the verification precision, where $m$ is the number of inputs that are proved to be robust, and $n$ is the number of all the inputs. The time is measured by seconds.
We will denote by RST and RST$^*$ respectively the verification precision of the original tool and our tool,
and by ACC the time reduction rate.

\begin{table}[t]
	\centering
	\caption{The verification results using {\sf MipVerify} and {\sf MipVerify}$^*$}
	\scalebox{0.9}{\subtable[The result of FNN\_1]{
			\begin{tabular}{|C{0.8cm}|R{1.4cm}|R{1.8cm}|R{2cm}|R{1.5cm}|R{1.5cm}|R{1.5cm}|}
				\hline
				& \multicolumn{6}{c|}{FNN\_1: $\langle$784,24,24,10$\rangle$, valid input:  99/100} \\ \hline
				$\epsilon$ &\centering  $t_{\mathit{sort}}$(s) & {\sf MipVerify}(s) &\centering ${\sf MipVerify}^*$(s) &\centering  ACC(\%) &\centering  RST &\centering  RST$^*$ \tabularnewline \hline
				1  & 0.15 & 84.20   & 2.18    & 97.21 & 96/100  & 96/100 \\ \hline
				3  & 0.20 & 275.48  & 70.82   & 74.22 & 93/100  & 93/100 \\ \hline
				5  & 0.37 & 526.75  & 348.12  & 33.84 & 85/100  & 85/100 \\ \hline
				7  & 0.44 & 764.89  & 692.29  & 9.43  & 53/100  & 53/100 \\ \hline
				9  & 0.49 & 996.45  & 971.56  & 2.45  & 37/100  & 37/100 \\ \hline
				11 & 0.51 & 1217.51 & 1210.78 & 0.51  & 19/100  & 19/100 \\ \hline
			\end{tabular}
			\label{tab:mipfnn1}
	}}
	
	\scalebox{0.9}{	\subtable[The result of FNN\_2]{
			\begin{tabular}{|C{0.8cm}|R{1.4cm}|R{1.8cm}|R{2cm}|R{1.5cm}|R{1.5cm}|R{1.5cm}|}
				\hline
				& \multicolumn{6}{c|}{FNN\_2 $\langle$784,100,100,10$\rangle$ valid input:  98/100} \\ \hline
				$\epsilon$ &\centering  $t_{\mathit{sort}}$(s) & {\sf MipVerify}(s) &\centering ${\sf MipVerify}^*$(s) &\centering  ACC(\%) &\centering  RST &\centering  RST$^*$ \tabularnewline \hline
				1  & 0.92 & 1033.85 & 60.65    & 94.04  & 98/100  & 98/100 \\ \hline
				3  & 1.18 & 3400.88 & 1222.44  & 64.01  & 96/100  & 96/100 \\ \hline
				5  & 1.36 & 8301.88 & 7635.60  & 8.01   & 93/100  & 93/100 \\ \hline
				7  & 1.93 & --      & --       & --     & --      & --     \\ \hline
				9  & 1.86 & --      & --       & --     & --      & --     \\ \hline
				11 & 2.20 & --      & --       & --     & --      & --     \\ \hline
			\end{tabular}
			\label{tab:mipfnn2}
	}}
	
	\scalebox{0.9}{	\subtable[The result of FNN\_3]{
			\begin{tabular}{|C{0.8cm}|R{1.4cm}|R{1.8cm}|R{2cm}|R{1.5cm}|R{1.5cm}|R{1.5cm}|}
				\hline
				& \multicolumn{6}{c|}{FNN\_3 $\langle$784,100,100,100,100,100,10$\rangle$ valid input:  99/100} \\ \hline
				$\epsilon$ &\centering  $t_{\mathit{sort}}$(s) & {\sf MipVerify}(s) &\centering ${\sf MipVerify}^*$(s) &\centering  ACC(\%) &\centering  RST &\centering  RST$^*$ \tabularnewline \hline
				1  & 2.37 & 3802.77 & 2741.02  & 27.86  & 99/100  & 99/100 \\ \hline
				3  & 3.68 & --      & --       & --     & --      & --     \\ \hline
				5  & 4.81 & --      & --       & --     & --      & --     \\ \hline
				7  & 5.23 & --      & --       & --     & --      & --     \\ \hline
				9  & 5.41 & --      & --       & --     & --      & --     \\ \hline
				11 & 6.13 & --      & --       & --     & --      & --     \\ \hline
			\end{tabular}
			\label{tab:mipfnn3}
	}}
	\label{tab:mipfnn}
	\vspace{-5mm}
\end{table}

\vspace{-5mm}
\subsubsection{Performance on {\sf MipVerify}.}
Table~\ref{tab:mipfnn} shows the verification results using {\sf MipVerify} and ${\sf MipVerify}^*$ on the three neural networks.
One can see that both tools return the same verification results under different perturbation distance thresholds and networks.
${\sf MipVerify}^*$ is more efficient than ${\sf MipVerify}$ in all the cases.
There is a significant speedup  when the perturbation threshold is small, \eg,  $\epsilon\leq 3$. It can even achieve $97\%$ time reduction for small neural network when $\epsilon=1$.
We can observe that the sorting time is almost linear with a very small coefficient \eg, 0.04 for FNN\_1 and 0.37 for FNN\_3, respectively.

The reason of the acceleration is that ${\sf MipVerify}^*$ will not spend extra time on verifying those labels that are proved impossible 
to be misclassified to or the robustness is already falsified. 
In our approach, we exclude such impossible labels during sorting as described in Section \ref{subsec:loc}. In contrast, {\sf MipVerify} treats all the classified labels except the original 
one with no difference and tries to find an adversarial example within a perturbation threshold for all labels, which incurs more time on finding solutions.

The experimental results also show that the acceleration of verification decreases
with the increasing of perturbation distance threshold and the size of network.
This is because that the increase of perturbation distance threshold will make more labels to be target labels to which adversarial examples exist.
In the worst case, our approach will not accelerate the verification due to the intrinsic NP-completeness of the problem.
As shown in Table~\ref{tab:mipfnn2} and~\ref{tab:mipfnn3},
 both tools run out of time when $\epsilon$ is too lager for the networks FNN\_2 and FNN\_3.

%

\begin{table}[t]
	\centering
	\caption{The verification results using {\sf DeepZ} and {\sf DeepZ}$^*$}  
	\scalebox{0.9}{\subtable[The result of FNN\_1]{
			\begin{tabular}{|C{1cm}|R{1.5cm}|R{1.8cm}|R{2cm}|R{1.5cm}|R{1.5cm}|R{1.5cm}|}
				\hline
				& \multicolumn{6}{c|}{FNN\_1 $\langle$784,24,24,10$\rangle$ valid input: 99/100} \\ \hline
				$\epsilon$ &\centering  $t_{\mathit{sort}}$(s) &\centering  {\sf DeepZ}(s) &\centering  {\sf DeepZ}$^*$(s) &\centering  ACC(\%) &\centering  RST &\centering  RST$^*$ \tabularnewline \hline
				1  & 0.15 &4.01  &0.49  &83.80  & 96/100  & 96/100 \\ \hline
				3  & 0.20 &3.98  &1.44  &58.76  & 90/100  & 90/100 \\ \hline
				5  & 0.37 &4.04  &3.11  &13.61  & 54/100  & 54/100 \\ \hline
				7  & 0.44 &4.14  &4.30  &-14.55 & 24/100  & 24/100 \\ \hline
				9  & 0.49 &4.82  &4.30  &0.56   & 6/100   & 6/100  \\ \hline
				11 & 0.51 &4.11  &4.17  &-13.86 & 0/100   & 0/100  \\ \hline
			\end{tabular}
			\label{tab:deepzfnn1}
	}}

	\scalebox{0.9}{\subtable[The result of FNN\_2]{
			\begin{tabular}{|C{1cm}|R{1.5cm}|R{1.8cm}|R{2cm}|R{1.5cm}|R{1.5cm}|R{1.5cm}|}
				\hline
				& \multicolumn{6}{c|}{FNN\_2 $\langle$784,100,100,10$\rangle$ valid input: 98/100} \\ \hline
				$\epsilon$ &\centering  $t_{\mathit{sort}}$(s) &\centering  {\sf DeepZ}(s) &\centering  {\sf DeepZ}$^*$(s) &\centering  ACC(\%) &\centering  RST &\centering  RST$^*$ \tabularnewline \hline
				1  & 0.92 & 41.29   & 1.92    & 93.10  & 97/100  & 97/100 \\ \hline
				3  & 1.18 & 42.14   & 15.37   & 60.71  & 89/100  & 89/100 \\ \hline
				5  & 1.36 & 43.12   & 40.52   & 2.88   & 56/100  & 56/100 \\ \hline
				7  & 1.93 & 44.05   & 43.75   & -3.70  & 13/100  & 13/100 \\ \hline
				9  & 1.86 & 44.78   & 44.95   & -4.54  & 2/100   & 2/100  \\ \hline
				11 & 2.20 & 45.56   & 45.76   & -5.27  & 0/100   & 0/100  \\ \hline
			\end{tabular}
			\label{tab:deepzfnn2}
	}}
	
	\scalebox{0.9}{\subtable[The result of FNN\_3]{
			\begin{tabular}{|C{1cm}|R{1.5cm}|R{1.8cm}|R{2cm}|R{1.5cm}|R{1.5cm}|R{1.5cm}|}
				\hline
				& \multicolumn{6}{c|}{FNN\_3 $\langle$784,100,100,100,100,100,10$\rangle$ valid input: 99/100} \\ \hline
				$\epsilon$ &\centering  $t_{\mathit{sort}}$(s) &\centering  {\sf DeepZ}(s) &\centering  {\sf DeepZ}$^*$(s) &\centering  ACC(\%) &\centering  RST &\centering  RST$^*$ \tabularnewline \hline
				1  & 2.37 & 94.69   & 52.52   & 42.03  & 99/100  & 99/100 \\ \hline
				3  & 3.68 & 102.78  & 92.23   & 6.69   & 81/100  & 81/100 \\ \hline
				5  & 4.81 & 117.45  & 94.58   & 15.37   & 33/100  & 33/100 \\ \hline
				7  & 5.23 & 127.98  & 104.20  & 14.49  & 3/100   & 3/100  \\ \hline
				9  & 5.41 & 132.84  & 123.96  & 2.61   & 0/100   & 0/100  \\ \hline
				11 & 6.13 & 137.91  & 121.01  & 7.80   & 0/100   & 0/100  \\ \hline
			\end{tabular}
			\label{tab:deepzfnn3}
	}}
	\label{deepzfnn}
	\vspace{-10mm}
\end{table}

\subsubsection{Performance on {\sf DeepZ}.}
Table~\ref{deepzfnn} reports the verification results using {\sf DeepZ} and {\sf DeepZ}$^*$.
The verification results of both tools are the same.
Because {\sf DeepZ} is an abstract interpretation based tool,
it is not surprised that
{\sf DeepZ} is more efficient than {\sf MipVerify}.
However, {\sf DeepZ} does not preserve completeness after introducing abstraction and therefore it may not be able to certify an input even if the neural network is robust to it against the preset perturbation range.
Our results confirm this.

\begin{table}[t]
	\centering
	\caption{The verification results using {\sf Neurify} and {\sf Neurify}$^*$}    
	\scalebox{0.9}{\subtable[The result of FNN\_1]{
			\begin{tabular}{|C{1cm}|R{1.5cm}|R{1.8cm}|R{2cm}|R{1.5cm}|R{1.5cm}|R{1.5cm}|}
				\hline
				& \multicolumn{6}{c|}{FNN\_1 $\langle$784,24,24,10$\rangle$ valid input: 99/100} \\ \hline
				$\epsilon$ &\centering $t_{\mathit{sort}}$(s) & {\sf Neurify}(s) &\centering  {\sf Neurify}$^*$(s) &\centering  ACC(\%) &\centering  RST &\centering  RST$^*$ \tabularnewline \hline
				1  & 0.15 &2.72   &0.37   &80.48  & 96/100  & 96/100 \\ \hline
				3  & 0.20 &11.96  &3.02   &73.03  & 93/100  & 93/100 \\ \hline
				5  & 0.37 &57.04  &53.45  &5.64   &\textbf{ 78}/100  & \textbf{79}/100 \\ \hline
				7  & 0.44 &73.71  &64.25  &12.23  &\textbf{ 44}/100 & \textbf{49}/100 \\ \hline
				9  & 0.49 &92.90  &88.38  &4.33   &\textbf{ 25}/100  & \textbf{30}/100 \\ \hline
				11 & 0.51 &86.19  &68.94  &19.41  & \textbf{8}/100   & \textbf{10}/100 \\ \hline
			\end{tabular}
			\label{tab:neurifyfnn1}
	}}
	\scalebox{0.9}{ 	\subtable[The result of FNN\_2]{
			\begin{tabular}{|C{1cm}|R{1.5cm}|R{1.8cm}|R{2cm}|R{1.5cm}|R{1.5cm}|R{1.5cm}|}
				\hline
				& \multicolumn{6}{c|}{FNN\_1 $\langle$784,100,100,10$\rangle$ valid input: 98/100} \\ \hline
				$\epsilon$ &\centering $t_{\mathit{sort}}$(s) & {\sf Neurify}(s) &\centering  {\sf Neurify}$^*$(s) &\centering  ACC(\%) &\centering  RST &\centering  RST$^*$ \tabularnewline \hline
				1  & 0.92 &3.09    &1.05    &35.93   & \textcolor{red}{100}/100 & \textcolor{red}{98}/100 \\ \hline
				3  & 1.18 &12.49   &13.70   &-19.23  & \textbf{93}/100  & \textbf{95}/100 \\ \hline
				5  & 1.36 &56.88   &61.92   &-11.24  & \textbf{68}/100  & \textbf{69}/100 \\ \hline
				7  & 1.93 &151.25  &166.75  &-11.53  & \textbf{35}/100  & \textbf{36}/100 \\ \hline
				9  & 1.86 &140.95  &170.09  &-21.99  & 6/100   & 6/100  \\ \hline
				11 & 2.20 &115.57  &128.02  &-12.69  & 1/100   & 1/100  \\ \hline
			\end{tabular}
			\label{tab:neurifyfnn2}
	}} 
	\scalebox{0.9}{	\subtable[The result of FNN\_3]{
			\begin{tabular}{|C{1cm}|R{1.5cm}|R{1.8cm}|R{2cm}|R{1.5cm}|R{1.5cm}|R{1.5cm}|}
				\hline
				& \multicolumn{6}{c|}{FNN\_3 $\langle$784,100,100,100,100,100,10$\rangle$ valid input: 99/100} \\ \hline
				$\epsilon$ &\centering $t_{\mathit{sort}}$(s) & {\sf Neurify}(s) &\centering  {\sf Neurify}$^*$(s) &\centering  ACC(\%) &\centering  RST &\centering  RST$^*$ \tabularnewline \hline
				1  & 2.37 & 5.42   & 2.73    & 5.89   & \textcolor{red}{100}/100 & \textcolor{red}{99}/100   \\ \hline
				3  & 3.68 & 52.76  & 47.90   & 2.23   & \textbf{87}/100  & \textbf{89}/100   \\ \hline
				5  & 4.81 & 116.79 & 114.04  & -1.77  & 43/100  & 43/100   \\ \hline
				7  & 5.23 & 122.05 & 146.27  & -24.13 & \textbf{7}/100   & \textbf{8}/100    \\ \hline
				9  & 5.41 & 131.22 & 146.91  & -16.08 & 1/100   & 1/100    \\ \hline
				11 & 6.13 & 120.69 & 139.89  & -20.99 & 0/100   & 0/100    \\ \hline
			\end{tabular}
			\label{tab:neurifyfnn3}
	}}
	\label{neurifyfnn}
	\vspace{-7mm}
\end{table}

The experimental results also show that when $\epsilon\leq 5$, our approach can accelerate the verification and improve the time by up to 93.10\% in some cases.
However, we also notice that when $\epsilon$ becomes larger, \eg, greater than $5$, the acceleration becomes weak, and it can be even slower than the original tool.
The reason is similar to the one of {\sf MipVerify}, \ie,
there are more target labels with the increasing of $\epsilon$. It is worth mentioning that
the reduced time is not always strictly monotonic. Tables~\ref{tab:deepzfnn2} and~\ref{tab:deepzfnn3} consists of such cases.
It would be interesting to perform an in-depth analysis of these cases.
One possible reason is that verification per target label cannot reuse the intermediate results of previous verifications.
One may use incremental verification approach to solve this problem.
We retain them as future work.

\vspace{-5mm}
\subsubsection{Performance on {\sf Neurify}.}
Table~\ref{neurifyfnn} shows the verification results using {\sf Neurify} and {\sf Neurify}$^*$.
Different from the results of the above two sections,
{\sf Neurify}$^*$  may produce different verification results from the original tool {\sf Neurify}.
For instance, {\sf Neurify}$^*$ finds more inputs to which FNN\_1 is robust using less time than {\sf Neurify} when $\epsilon\geq 5$ (cf. Table~\ref{tab:neurifyfnn1}).
That is because that {\sf Neurify} uses abstract and refinement in their verification approach,
and a maximum number of refinements is predefined to guarantee the termination. It returns \emph{unknown} once it exceeds the refinement threshold.
The order of target labels determines how fast that the tool reaches the threshold.
The experimental results show that by sorting the target labels we can reduce the possibility of reaching the refinement threshold.
Although our tool may take more verification time,
some inputs whose verification results are \emph{unknown} produced by the original tool
can be resolved as robust by our tool.

It is worth mentioning that {\sf Neurify} does not check whether an input can be correctly classified before verification, and regards it as robust even if it is always classified to an incorrect label.
Such cases are excluded in our algorithm, and therefore the number of safe inputs verified by {\sf Neurify}$^*$ may be less than the one verified by {\sf Neurify}, as shown by the red numbers in Table~\ref{neurifyfnn}.

%% file: conc.tex
\section{Related Work}\label{sec:relatedwork}
We discuss existing formal verification techniques for neural networks (cf.~\cite{LiuALBK19,HKRSSTWY19} for survey).
Neural network testing (e.g.,~\cite{SZSBEGF14,PCYJ17,MJZSXLCSLLZW18,PMGJCS17,IEAL18,CLCYZH18,TTCLZYC18,BHLS17,KGB17,BRB18,LCFSL20,CCFDZSL20,DZBS19} to cite a few)
are excluded, which are computationally less expensive and  are able to work with large networks, but at the cost of the provable
guarantees. 

Existing formal verification techniques can be broadly classified as either complete or incomplete ones. 
Complete techniques are based on constraint solvers such as SAT/SMT/MILP solving~\cite{ehlers2017formal,ehlers2017formal,katz2017reluplex,tjeng2017verifying} or refinement~\cite{wang2018efficient}. 
They do not produce false positives but have limited scalability and hardly handle neural networks containing more than a few hundreds of hidden units. 
In contrast, incomplete ones usually rely on approximation for better scalability, but they may produce false positives.
Such techniques mainly include duality~\cite{dvijotham2018dual,raghunathan2018certified,wong2017provable}, layer-by-layer approximations of the adversarial polytope~\cite{xiang2018output}, discretizing the search space~\cite{huang2017safety}, abstract interpretation~\cite{gehr2018ai2,singh2018fast,singh2019abstract}, linear approximations~\cite{weng2018towards,wong2017provable}, bounding the local Lipschitz constant~\cite{weng2018towards}, or bounding the activation of the ReLU with linear functions~\cite{weng2018towards}.

The complete robustness verification of ReLU-based neural network is essentially a collection of linear programming problems. For a neuron with a ReLU activation function, the function can be active or inactive, depending on the input.
Thus, every neuron is transformed into linear constraints. Consequently, the size of linear programming problems to solve increases exponentially with the number of neurons in a network, which is obviously not scalable.
Katz \etal\ proved that the robustness verification problem of DNNs is NP-complete~\cite{katz2017reluplex},
 which illustrates the necessity of devising algorithmically efficient verification method. They extended the classical simplex algorithm to solve this problem~\cite{katz2017reluplex}. However, the algorithm is still limited to small-scale neural networks. For example, verifying a feed-forward network with 5 inputs, 5 outputs and 300 total hidden neurons on a single data point can take a few hours~\cite{katz2017reluplex}.
Another solver-based verification system is Planet~\cite{ehlers2017formal}, which resorts to satisfiability (SAT) solvers. Although an adversarial example found by such approaches is a genuine one, their scalability is always an obstacle which prevents them from being applied to relatively large neural networks.


Incomplete verification techniques do not intend to solve the verification task directly. Instead, they tune the verification problem into a classical linear programming problem by over-approximating the adversarial polytope or the space of outputs of a network for a region of possible inputs. For instance,~\cite{wong2017provable} and~\cite{raghunathan2018certified} transform the verification problem into a convex optimization problem using relaxations to over-approximate the outputs of ReLU nodes. Another typical work~\cite{gehr2018ai2} leverages zonotopes for approximating each ReLU outputs. Dvijotham \etal\  propose to transform the verification problem into an unconstrained dual formulation using Lagrange relaxation and use gradient-descent to solve the optimization problem~\cite{dvijotham2018dual}.
Such over-approximation drastically improves the efficiency of obtaining provable adversarial accuracy results. However, incomplete verification may produce false positives. Recently, two novel abstraction-based frameworks for neural
network verification have been proposed~\cite{EGK20,AHKM20} which merge
several neurons into a single neuron and obtain a smaller, abstracted neural network, while prior work abstracts
the transformation of each neuron.

Our approach is orthogonal to, and can be integrated with
existing neural network verification techniques and abstraction-based frameworks, to accelerate robustness verification. Although both the symbolic interval analysis and linear relaxation techniques have been used in existing works,
to our knowledge, they are the first time used for ranking labels. Furthermore, our robustness verification methodology that reduces to
the verification for each target label and the verification approach for one target label are new to some extent.


\section{Conclusion and Future Work}\label{sec:conc}

In this work, we proposed a novel, generic approach to accelerate the robustness verification of DNNs.
The novelty of our approach is threefold. First, we showed that the overall verification problem can be reduced to the verification problem
for each target label. Second, we presented an efficient and effective approach for ranking labels.
Finally, we integrated our approach into three recent promising DNN verification tools.
Experimental results showed that our approach is effective when the perturbation distance is set in a
reasonable range. 

In future, we plan to investigate incremental verification approaches so that the intermediate results of previous verifications
could be reused for verifying later labels.
We also plan to verify industry-level networks using more powerful hardware such as GPU. We believe that the improvement in efficiency makes it possible to verify DNN-based systems which is crucial to apply to safety-critical domains.

%% file: main.bbl
\begin{thebibliography}{10}
\providecommand{\url}[1]{\texttt{#1}}
\providecommand{\urlprefix}{URL }
\providecommand{\doi}[1]{https://doi.org/#1}

\bibitem{Apollo}
Apollo: An open, reliable and secure software platform for autonomous driving
  systems. \url{http://apollo.auto} (2018)

\bibitem{AHKM20}
Ashok, P., Hashemi, V., Kret{\'{\i}}nsk{\'{y}}, J., Mohr, S.: Deepabstract:
  Neural network abstraction for accelerating verification. In: Proceedings of
  The 18th International Symposium on Automated Technology for Verification and
  Analysis {(ATVA)} (2020)

\bibitem{bezanson2017julia}
Bezanson, J., Edelman, A., Karpinski, S., Shah, V.B.: Julia: A fresh approach
  to numerical computing. SIAM review  \textbf{59}(1),  65--98 (2017)

\bibitem{BHLS17}
Bhagoji, A.N., He, W., Li, B., Song, D.: Exploring the space of black-box
  attacks on deep neural networks. CoRR  \textbf{abs/1712.09491} (2017)

\bibitem{BRB18}
Brendel, W., Rauber, J., Bethge, M.: Decision-based adversarial attacks:
  Reliable attacks against black-box machine learning models. In: Proceedings
  of the International Conference on Learning Representations (ICLR) (2018)

\bibitem{carlini2017towards}
Carlini, N., Wagner, D.: Towards evaluating the robustness of neural networks.
  In: 2017 IEEE Symposium on Security and Privacy (SP). pp. 39--57. IEEE (2017)

\bibitem{CCFDZSL20}
Chen, G., Chen, S., Fan, L., Du, X., Zhao, Z., Song, F., Liu, Y.: Who is real
  bob? adversarial attacks on speaker recognition systems. CoRR
  \textbf{abs/1911.01840} (2019)

\bibitem{cheng2017maximum}
Cheng, C.H., N{\"u}hrenberg, G., Ruess, H.: Maximum resilience of artificial
  neural networks. In: International Symposium on Automated Technology for
  Verification and Analysis. pp. 251--268. Springer (2017)

\bibitem{CLCYZH18}
Cheng, M., Le, T., Chen, P., Zhang, H., Yi, J., Hsieh, C.: Query-efficient
  hard-label black-box attack: An optimization-based approach. In: Proceedings
  of the 7th International Conference on Learning Representations (ICLR) (2019)

\bibitem{CGGS12}
Ciresan, D.C., Giusti, A., Gambardella, L.M., Schmidhuber, J.: Deep neural
  networks segment neuronal membranes in electron microscopy images. In:
  Proceedings of the 26th Annual Conference on Neural Information Processing
  Systems. pp. 2852--2860 (2012)

\bibitem{DZBS19}
Duan, Y., Zhao, Z., Bu, L., Song, F.: Things you may not know about adversarial
  example: {A} black-box adversarial image attack. CoRR
  \textbf{abs/1905.07672} (2019)

\bibitem{dvijotham2018dual}
Dvijotham, K., Stanforth, R., Gowal, S., Mann, T.A., Kohli, P.: A dual approach
  to scalable verification of deep networks. In: UAI. pp. 550--559 (2018)

\bibitem{ehlers2017formal}
Ehlers, R.: Formal verification of piece-wise linear feed-forward neural
  networks. In: International Symposium on Automated Technology for
  Verification and Analysis. pp. 269--286. Springer (2017)

\bibitem{EGK20}
Elboher, Y.Y., Gottschlich, J., Katz, G.: An abstraction-based framework for
  neural network verification. In: Proceedings of The 32nd International
  Conference on Computer-Aided Verification (CAV) (2020)

\bibitem{fischetti2017deep}
Fischetti, M., Jo, J.: Deep neural networks as 0-1 mixed integer linear
  programs: A feasibility study. arXiv preprint arXiv:1712.06174  (2017)

\bibitem{gehr2018ai2}
Gehr, T., Mirman, M., Drachsler-Cohen, D., Tsankov, P., Chaudhuri, S., Vechev,
  M.: Ai2: Safety and robustness certification of neural networks with abstract
  interpretation. In: 2018 IEEE Symposium on Security and Privacy (SP). pp.
  3--18. IEEE (2018)

\bibitem{goodfellow2014explaining}
Goodfellow, I.J., Shlens, J., Szegedy, C.: Explaining and harnessing
  adversarial examples. arXiv preprint arXiv:1412.6572  (2014)

\bibitem{he2016deep}
He, K., Zhang, X., Ren, S., Sun, J.: Deep residual learning for image
  recognition. In: Proceedings of the IEEE conference on computer vision and
  pattern recognition. pp. 770--778 (2016)

\bibitem{hein2017formal}
Hein, M., Andriushchenko, M.: Formal guarantees on the robustness of a
  classifier against adversarial manipulation. In: Advances in Neural
  Information Processing Systems. pp. 2266--2276 (2017)

\bibitem{hinton2012deep}
Hinton, G., Deng, L., Yu, D., Dahl, G., Mohamed, A.r., Jaitly, N., Senior, A.,
  Vanhoucke, V., Nguyen, P., Kingsbury, B., et~al.: Deep neural networks for
  acoustic modeling in speech recognition. IEEE Signal processing magazine
  \textbf{29} (2012)

\bibitem{Holley18}
Holley, P.: Texas becomes the latest state to get a self-driving car service.
  \url{https://www.washingtonpost.com/news/innovations/wp/2018/05/07/texas-becomes-the-latest-state-to-get-a-self-driving-car-service}
  (May 2018)

\bibitem{huang2017densely}
Huang, G., Liu, Z., Van Der~Maaten, L., Weinberger, K.Q.: Densely connected
  convolutional networks. In: Proceedings of the IEEE conference on computer
  vision and pattern recognition. pp. 4700--4708 (2017)

\bibitem{HKRSSTWY19}
Huang, X., Kroening, D., Ruan, W., Sharp, J., Sun, Y., Thamo, E., Wu, M., Yi,
  X.: Safety and trustworthiness of deep neural networks: {A} survey. CoRR
  \textbf{abs/1812.08342v4} (2019), \url{http://arxiv.org/abs/1812.08342v4}

\bibitem{huang2017safety}
Huang, X., Kwiatkowska, M., Wang, S., Wu, M.: Safety verification of deep
  neural networks. In: International Conference on Computer Aided Verification.
  pp. 3--29. Springer (2017)

\bibitem{IEAL18}
Ilyas, A., Engstrom, L., Athalye, A., Lin, J.: Black-box adversarial attacks
  with limited queries and information. In: Proceedings of the 35th
  International Conference on Machine Learning (ICML). pp. 2142--2151 (2018)

\bibitem{katz2017reluplex}
Katz, G., Barrett, C., Dill, D.L., Julian, K., Kochenderfer, M.J.: Reluplex: An
  efficient smt solver for verifying deep neural networks. In: International
  Conference on Computer Aided Verification. pp. 97--117. Springer (2017)

\bibitem{krizhevsky2012imagenet}
Krizhevsky, A., Sutskever, I., Hinton, G.E.: Imagenet classification with deep
  convolutional neural networks. In: Advances in neural information processing
  systems. pp. 1097--1105 (2012)

\bibitem{KGB17}
Kurakin, A., Goodfellow, I., Bengio, S.: Adversarial examples in the physical
  world. In: Proceedings of the International Conference on Learning
  Representations (ICLR) (2017)

\bibitem{lecun1998gradient}
LeCun, Y., Bottou, L., Bengio, Y., Haffner, P., et~al.: Gradient-based learning
  applied to document recognition. Proceedings of the IEEE  \textbf{86}(11),
  2278--2324 (1998)

\bibitem{LCFSL20}
Lei, Y., Chen, S., Fan, L., Song, F., Liu, Y.: Advanced evasion attacks and
  mitigations on practical ml-based phishing website classifiers. CoRR
  \textbf{abs/2004.06954} (2020)

\bibitem{LiuALBK19}
Liu, C., Arnon, T., Lazarus, C., Barrett, C.W., Kochenderfer, M.J.: Algorithms
  for verifying deep neural networks. CoRR  \textbf{abs/1903.06758} (2019)

\bibitem{lomuscio2017approach}
Lomuscio, A., Maganti, L.: An approach to reachability analysis for
  feed-forward relu neural networks. arXiv preprint arXiv:1706.07351  (2017)

\bibitem{MJZSXLCSLLZW18}
Ma, L., Juefei{-}Xu, F., Zhang, F., Sun, J., Xue, M., Li, B., Chen, C., Su, T.,
  Li, L., Liu, Y., Zhao, J., Wang, Y.: Deepgauge: multi-granularity testing
  criteria for deep learning systems. In: Proceedings of the 33rd {ACM/IEEE}
  International Conference on Automated Software Engineering (ASE). pp.
  120--131 (2018)

\bibitem{moore2009introduction}
Moore, R.E., Kearfott, R.B., Cloud, M.J.: Introduction to interval analysis,
  vol.~110. Siam (2009)

\bibitem{moosavi2016deepfool}
Moosavi-Dezfooli, S.M., Fawzi, A., Frossard, P.: Deepfool: a simple and
  accurate method to fool deep neural networks. In: Proceedings of the IEEE
  conference on computer vision and pattern recognition. pp. 2574--2582 (2016)

\bibitem{PMGJCS17}
Papernot, N., McDaniel, P.D., Goodfellow, I.J., Jha, S., Celik, Z.B., Swami,
  A.: Practical black-box attacks against machine learning. In: Proceedings of
  the {ACM} on Asia Conference on Computer and Communications Security
  (AsiaCCS). pp. 506--519 (2017)

\bibitem{PCG15}
Parag, T., Ciresan, D.C., Giusti, A.: Efficient classifier training to minimize
  false merges in electron microscopy segmentation. In: Proceedings of 2015
  {IEEE} International Conference on Computer Vision. pp. 657--665 (2015)

\bibitem{peck2017lower}
Peck, J., Roels, J., Goossens, B., Saeys, Y.: Lower bounds on the robustness to
  adversarial perturbations. In: Advances in Neural Information Processing
  Systems. pp. 804--813 (2017)

\bibitem{PCYJ17}
Pei, K., Cao, Y., Yang, J., Jana, S.: Deepxplore: Automated whitebox testing of
  deep learning systems. In: Proceedings of the 26th Symposium on Operating
  Systems Principles (SOSP). pp. 1--18 (2017)

\bibitem{raghunathan2018certified}
Raghunathan, A., Steinhardt, J., Liang, P.: Certified defenses against
  adversarial examples. arXiv preprint arXiv:1801.09344  (2018)

\bibitem{SWS17}
Shen, D., Wu, G., , Suk, H.I.: Deep learning in medical image analysis. Annual
  Review of Biomedical Engineering  \textbf{19},  221--248 (2017)

\bibitem{singh2018fast}
Singh, G., Gehr, T., Mirman, M., P{\"u}schel, M., Vechev, M.: Fast and
  effective robustness certification. In: Advances in Neural Information
  Processing Systems. pp. 10802--10813 (2018)

\bibitem{singh2019abstract}
Singh, G., Gehr, T., P{\"u}schel, M., Vechev, M.: An abstract domain for
  certifying neural networks. Proceedings of the ACM on Programming Languages
  \textbf{3}(POPL), ~41 (2019)

\bibitem{singh2018boosting}
Singh, G., Gehr, T., P{\"{u}}schel, M., Vechev, M.T.: Boosting robustness
  certification of neural networks. In: 7th International Conference on
  Learning Representations ({ICLR}) (2019)

\bibitem{stewart2018tesla}
Stewart, J.: Tesla's autopilot was involved in another deadly car crash.
  \url{https://www.wired.com/story/tesla-autopilot-self-driving-crash-california/}
  (2018)

\bibitem{szegedy2016rethinking}
Szegedy, C., Vanhoucke, V., Ioffe, S., Shlens, J., Wojna, Z.: Rethinking the
  inception architecture for computer vision. In: Proceedings of the IEEE
  conference on computer vision and pattern recognition. pp. 2818--2826 (2016)

\bibitem{szegedy2013intriguing}
Szegedy, C., Zaremba, W., Sutskever, I., Bruna, J., Erhan, D., Goodfellow, I.,
  Fergus, R.: Intriguing properties of neural networks. In: Proceedings of
  International Conference on Learning Representations (2014)

\bibitem{SZSBEGF14}
Szegedy, C., Zaremba, W., Sutskever, I., Bruna, J., Erhan, D., Goodfellow, I.,
  Fergus, R.: Intriguing properties of neural networks. In: Proceedings of the
  2nd International Conference on Learning Representations (ICLR) (2014)

\bibitem{tjeng2017verifying}
Tjeng, V., Tedrake, R.: Verifying neural networks with mixed integer
  programming. corr abs/1711.07356 (2017). arXiv preprint arXiv:1711.07356
  (2017)

\bibitem{tjeng2017evaluating}
Tjeng, V., Xiao, K.Y., Tedrake, R.: Evaluating robustness of neural networks
  with mixed integer programming. In: 7th International Conference on Learning
  Representations ({ICLR}) (2019)

\bibitem{TTCLZYC18}
Tu, C., Ting, P., Chen, P., Liu, S., Zhang, H., Yi, J., Hsieh, C., Cheng, S.:
  Autozoom: Autoencoder-based zeroth order optimization method for attacking
  black-box neural networks. In: Proceedings of the 33rd {AAAI} Conference on
  Artificial Intelligence (AAAI). pp. 742--749 (2019)

\bibitem{wang2018efficient}
Wang, S., Pei, K., Whitehouse, J., Yang, J., Jana, S.: Efficient formal safety
  analysis of neural networks. In: Advances in Neural Information Processing
  Systems. pp. 6367--6377 (2018)

\bibitem{WangPWYJ18}
Wang, S., Pei, K., Whitehouse, J., Yang, J., Jana, S.: Formal security analysis
  of neural networks using symbolic intervals. In: 27th {USENIX} Security
  Symposium. pp. 1599--1614 (2018)

\bibitem{Waymo}
Waymo: A self-driving technology development company. \url{https://waymo.com/}
  (2009)

\bibitem{weng2018towards}
Weng, T.W., Zhang, H., Chen, H., Song, Z., Hsieh, C.J., Boning, D., Dhillon,
  I.S., Daniel, L.: Towards fast computation of certified robustness for relu
  networks. arXiv preprint arXiv:1804.09699  (2018)

\bibitem{wong2017provable}
Wong, E., Kolter, J.Z.: Provable defenses against adversarial examples via the
  convex outer adversarial polytope. arXiv preprint arXiv:1711.00851  (2017)

\bibitem{xiang2018output}
Xiang, W., Tran, H.D., Johnson, T.T.: Output reachable set estimation and
  verification for multilayer neural networks. IEEE transactions on neural
  networks and learning systems  \textbf{29}(11),  5777--5783 (2018)

\bibitem{zhang2018efficient}
Zhang, H., Weng, T.W., Chen, P.Y., Hsieh, C.J., Daniel, L.: Efficient neural
  network robustness certification with general activation functions. In:
  Advances in neural information processing systems. pp. 4939--4948 (2018)

\end{thebibliography}
